\documentclass[letterpaper]{article} 
\usepackage{aaai25}  
\usepackage{times}  
\usepackage{helvet}  
\usepackage{courier}  
\usepackage[hyphens]{url}  
\usepackage{graphicx} 
\usepackage{natbib}  
\usepackage{caption} 
\usepackage{algorithm}
\usepackage{algorithmic}

\usepackage{newfloat}
\usepackage{listings}
\DeclareCaptionStyle{ruled}{labelfont=normalfont,labelsep=colon,strut=off} 
\floatstyle{ruled}
\newfloat{listing}{tb}{lst}{}
\floatname{listing}{Listing}
\title{title}

\title{Replacing Paths with Connection-Biased Attention for \\Knowledge Graph Completion}
\author {
    Sharmishtha Dutta\textsuperscript{\rm 1},
    Alex Gittens\textsuperscript{\rm 1},
    Mohammed J. Zaki\textsuperscript{\rm 1},
    Charu C. Aggarwal\textsuperscript{\rm 2}
}
\affiliations {
    \textsuperscript{\rm 1}Department of Computer Science, Rensselaer Polytechnic Institute, Troy, NY, USA\\
    \textsuperscript{\rm 2}IBM T. J. Watson Research Center, Yorktown Heights, NY, USA\\
    duttas@rpi.edu, gittea@rpi.edu, zaki@cs.rpi.edu, charu@us.ibm.com
}

\usepackage{amssymb, amsmath, multirow}
\begin{document}

\maketitle

\begin{abstract}
 Knowledge graph (KG) completion aims to identify additional facts that can be inferred from the existing facts in the KG. Recent developments in this field have explored this task in the inductive setting, where at test time one sees entities that were not present during training; the most performant models in the inductive setting have employed path encoding modules in addition to standard subgraph encoding modules. This work similarly focuses on KG completion in the inductive setting, without the explicit use of path encodings, which can be time-consuming and introduces several hyperparameters that require costly hyperparameter optimization. Our approach uses a Transformer-based subgraph encoding module only; we introduce connection-biased attention and entity role embeddings into the subgraph encoding module to eliminate the need for an expensive and time-consuming path encoding module. Evaluations on standard inductive KG completion benchmark datasets demonstrate that our \textbf{C}onnection-\textbf{B}iased \textbf{Li}nk \textbf{P}rediction (CBLiP) model has superior performance to models that do not use path information. Compared to models that utilize path information, CBLiP shows competitive or superior performance while being faster. Additionally, to show that the effectiveness of connection-biased attention and entity role embeddings also holds in the transductive setting, we compare CBLiP's performance on the relation prediction task in the transductive setting.
 
\end{abstract}

\section{Introduction}

Knowledge graphs (KGs) store facts expressed in the forms of relationships between entities. Each fact is represented as a triple (\textit{h,r,t}) or (\textit{head, relation, tail}). Here, the \textit{head} and \textit{tail} represent entities such as people, places, and institutions, and the \textit{relation} represents the relation between the two entities. These facts are modeled as a directed graph with labeled edges, where each entity is a vertex in the graph, the relations are the edge labels, and edges are directed from the head to the tail entities. 

KGs are often constructed by crowdsourced data, or by using off-the-shelf fact extraction tools. Therefore, in addition to containing spurious information, they can omit facts that are implicit in the observed data. These omissions hinder the usefulness of KGs in various downstream tasks such as question answering in search engines. This has motivated researchers to use statistical relational learning to infer missing entities or relations from incomplete triples. Earlier approaches to KG completion were developed for the transductive setting, where the triples to be completed consist of entities and relations seen during training. While this stream of research has yielded numerous high-performing models, these models cannot be used when the triples to be completed contain entities that were not seen at training time.

\begin{figure}[t]
    \centering
    \includegraphics[width=\linewidth]{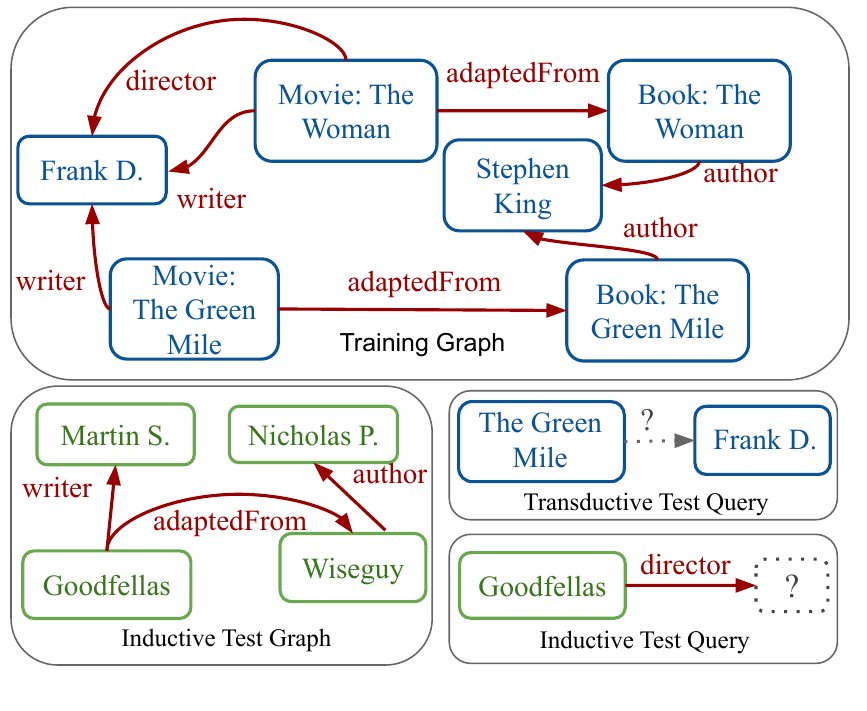}
    \caption{Example of training graph and test queries for KG completion in transductive and inductive settings. }
    \label{fig:motivating_example}
\end{figure}
Real-world KGs grow continuously as new facts are added, and the set of entities may grow over time, meaning that at any given point, completion models may need to be used on entities that were not seen during training. Inductive KG Completion assumes the relations between entities remain the same in the newly added facts. To illustrate this setting, Figure~\ref{fig:motivating_example} depicts a training graph that reflects the correlation between the role of a director and screenwriter. The same artist often plays these two roles when the screenplay is adapted from an existing novel. In the transductive KG completion setting, a relation prediction task involves queries where both entities are present in the training graph and the most plausible relation is to be determined. In the inductive setting, the test graph contains entities added to the graph after the model was trained, and the inductive KG completion task to find the best option for a missing entity involves entities unseen during training (marked by green ink in the figure).


Naturally, in the inductive setting, graph-learning-based approaches rely greatly on the information provided by neighbors of the incomplete triple of interest. This has led to several Graph Convolution Network (GCN) based models that encode the surrounding subgraph to learn about relation interactions. Paths between entities have also been utilized in several models \cite{lin2022incorporating, li2023inductive, pan2022inductive, zhu2021neural}, as explicitly encoding path information has resulted in a great performance boost. Here, path information is usually defined as an ordered sequence of relations between two entities. For example, two paths between \verb|Frank D.| and \verb|Stephen King|, according to Figure \ref{fig:motivating_example}, are $(writer^{-1}, adaptedFrom, author)$ and $(director^{-1}, adaptedFrom, author)$. 

The use of paths necessitates multiple decisions, such as determining the ideal path length to obtain meaningful information and exclude noise, the number of paths to be extracted, and various design choices for combining this information with the subgraph information. Additionally, path extraction between entities on the fly and representing them in the model adds overhead to the training time and parameter count.

Transformers have replaced Recurrent Neural Networks for sequence modeling precisely because appropriate positional encoding allows attention alone to suffice; we hypothesize that when represented as sequences of triples, appropriate positional encodings similarly unleash the full power of attention, obviating the need for explicit and costly modeling of path information. Indeed, by adapting the connection-biased attention from GRAN \cite{wang2021link}-- where it was used to better represent single $n$-ary facts in a knowledge base-- to the representation of subgraphs of a KG, we demonstrate that Transformers alone, without specialized submodules for path representation, suffice to perform accurate KG completion.

\noindent\textbf{Main Contributions.}
Our key contributions are:
\begin{enumerate}
    \item We introduce CBLiP: a context-aware Transformer-based model, with a novel connection-biased attention module at its core for reasoning in KGs.
    \item We introduce entity roles, a simple and effective construct to represent unseen entities in a subgraph, as an alternative to conventional relative distance-based entity labeling in the inductive link prediction setting.
    \item We demonstrate the effectiveness of CBLiP by comparing its performance on the entity prediction task in the inductive setting with that of state-of-the-art models on benchmark datasets and showing that it achieves best-performing or competitive results.
    \item We highlight CBLip's effectiveness across settings by similarly evaluating its performance on a transductive relation prediction task.
\end{enumerate}

\section{Related Work}
Knowledge graph completion is often based on learning continuous vector embedding of entities and relations. KG completion garnered attention in the early 2010s with TransE~\cite{bordes2013translating} where the distance between learned embedding vectors determined the plausibility of a triple. GraIL~\cite{teru2020inductive} introduced the inductive learning task and the standard benchmark datasets for this task and has garnered much attention as a more practical approach. In this section, we discuss the development of both settings. 

\subsection{Transductive Learning Models}
Initial work in the transductive setting focused on triple-based models, exploiting entities' inherent properties. Recent work has focused on gathering information about the neighborhood surrounding an entity.

\subsubsection{Translational Distance-based Models}

Research in translation-based models focused on learning embeddings that satisfy specific properties. For example, TransE \cite{bordes2013translating} models relations as translation vectors between head and tail entities, aiming to maintain \(\mathbf{h+r} \approx \mathbf{t}\). Despite its limitations in representing many-to-one and one-to-many relations, TransE remains a lightweight and straightforward model. Subsequent models like RotatE \cite{sun2018rotate} and QuatE \cite{zhang2019quaternion} embed entities in complex and quaternion spaces, respectively.
\subsubsection{Factorization Based Models}
Another family of models aims to capture semantic similarity by observing pairwise interactions between entities. RESCAL \cite{nickel2011three} was an early model with this idea. It was followed by a simplified variation DistMult \cite{yang2015embedding} and a complex number variation ComplEx \cite{trouillon2016complex}. SimplE \cite{kazemi2018simple} extends Canonical Polyadic (CP) decomposition by removing the independence between learned entity representations.

\subsubsection{Neural Models}

PathCon \cite{wang2021relational} utilizes a message-passing mechanism to aggregate edge-based local context and paths between node pairs for relation prediction. It does not learn any node embeddings, limiting its applicability for entity prediction.


\subsection{Inductive Learning Models}

Inductive KG completion aims to extend the task to unseen entities. Models first relied on Graph Neural Networks (GNNs) and more recently on Transformers to encode context information by aggregating neighborhood interactions. Many models have leveraged path information between the \textit{head} and \textit{tail} entities by incorporating a path encoding module in addition to the subgraph module that aggregates neighborhood interactions.

\subsubsection{Models without Path Information}
GraIL \cite{teru2020inductive} selects the common neighbors of the target head and tail entities as the context of a target triple to be scored. The model employs double-radius labeling (distance from head, distance from tail) to denote each entity's relative position and learns embeddings for relations through attention computation inside an Relational GCN module. The final score is a function of the triple in consideration and the encoded subgraph surrounding it. CoMPILE \cite{mai2021communicative} extends GraIL by considering directedness in its subgraph encoding module and computing edge (triple) attentions to learn edge embedding. TACT \cite{chen2021topology} expands GraIL by proposing an additional relation correlation module by learning 6 predefined categories of interactions via unique linear transformations for each kind. This categorization of TACT is closely related to our approach to constructing a connection-biased adjacency matrix of 7 categories. However, TACT incorporates this information into its subgraph encoding module whereas we use it to compute connection-biased attention in the transformer layers.


\subsubsection{Models with Path Information}
ConGLR \cite{lin2022incorporating} expands on Grail by modifying the subgraph encoding module. Additionally, it constructs a context graph that uses the relational paths involving the neighborhood entities. A combination function and a weighted aggregation function are employed to encode the paths represented as a sequence of relations and to combine the paths, respectively. The final scoring function integrates context information and path-based logical reasoning. Report \cite{li2023inductive} uses successive stacks of transformer layers for context encoding and path encoding. A hierarchical structure of transformer layers is utilized to fuse the query and the representations of context and path to compute a final score. \cite{pan2022inductive} proposed LogCo where a GCN-based subgraph module is complemented with a path encoding module. Each path representation is compared with the target relation to compute an attention score based on similarity. The model uses positive and negative path samples for a contrastive training regime. These models add the overhead of path representation and path aggregation during training and path extraction during training and inference time. 

NBFNet \cite{zhu2021neural} offers a more scalable solution to this by generalizing Bellman-Ford algorithm for finding the shortest paths. The model learns entity pair representations as well as path representations utilizing the distributive properties of the generalized operators. This allows parallel scoring of query triples that share the same entity-relation pairs and the models suffer from time complexity as well as memory overhead due to additional support needed to represent paths. While these models perform better than those without path information, they come with a drawback of real-time path extraction for inference triples. Since the core of inductive link prediction is to conduct reasoning in unseen entities during training, preprocessing of paths is not a realistic choice for inference triples.
\subsection{Models with Rule Extraction}

RPJE \cite{niu2020rule} combines rules and paths by injecting length-2 rules into KG embeddings. DRUM \cite{sadeghian2019drum} and NeuralLP \cite{yang2017differentiable} extract probabilistic first-order logic rules to assign weights to paths between entities. RuleN \cite{meilicke2018fine} assigns confidence to rules by randomized process. While these methods are efficient in learning short and simple rules, they suffer from scalability issues while finding frequent patterns in large graphs.

\section{Problem Formulation}

Relational data can be modeled as a directed heterogeneous graph $\mathcal{G}=(\mathcal{E},\mathcal{R}, \mathcal{F})$ where $\mathcal{E}$ and $\mathcal{R}$ represent the set of entities and relations modeled as nodes and edge types in the graph, respectively. $\mathcal{F}\subset \mathcal{E}\times\mathcal{R}\times\mathcal{E}$ represents the labeled edges or fact triples represented as ordered tuples of head-entity, relation, tail-entity. 

The inductive KG completion task comprises the following components.
\begin{enumerate}
    \item \textbf{Training graph}: $\mathcal{G}_{\mathrm{train}}=(\mathcal{E}_{\mathrm{train}},\mathcal{R}, \mathcal{F}_{\mathrm{train}})$ where $\mathcal{F}_{\mathrm{train}}\subset \mathcal{E}_{\mathrm{train}}\times\mathcal{R}\times\mathcal{E}_{\mathrm{train}}$ is the set of training facts.
    \item \textbf{Validation triples}:  $\mathcal{F}_{\mathrm{valid}}\subset\mathcal{E}_{\mathrm{train}}\times\mathcal{R}\times\mathcal{E}_{\mathrm{train}}$
    \item \textbf{Test graph}: $\mathcal{G}_{\mathrm{test}}=(\mathcal{E}_{\mathrm{test}},\mathcal{R}, \mathcal{F}_{\mathrm{test}})$ where $\mathcal{F}_{\mathrm{test}}\subset \mathcal{E}_{\mathrm{test}}\times\mathcal{R}\times\mathcal{E}_{\mathrm{test}}$. This serves as the fact graph for the test-time inference triples.
    \item \textbf{Test-time inference triples}: $\mathcal{F}_{\mathrm{infer}}\subset\mathcal{E}_{\mathrm{test}}\times\mathcal{R}\times\mathcal{E}_{\mathrm{test}}$. Given an incomplete fact from $\mathcal{F}_{\mathrm{infer}}$, the aim is to complete it using information from $\mathcal{G}_{\mathrm{test}}$ and the model trained on $\mathcal{G}_{\mathrm{train}}$.
\end{enumerate}

The inductive setting is characterized by the fact that $\mathcal{E}_{\mathrm{train}}\cap\mathcal{E}_{\mathrm{test}}=\varnothing$,
i.e., the entities seen at test time were unseen at training time.

Our goal is to find a model that computes a score  $s$ for a triple $\langle h,r,t\rangle$. We hypothesize that a plausibility score of a triple can be determined using information present in the ego graphs of the \textit{head} and \textit{tail} entities. The $k$-hop ego graph $\mathcal{N}_{e}$ of entity $e$ consists of the triples in its $k$-hop enclosing subgraph. Thus, we model the score with 
\begin{equation}
    s = g(h, r, t, \mathcal{N}_h, \mathcal{N}_t)
    \label{eqn:ind_model}
\end{equation}
Here, $g$ is a function of the triple and its local contexts; in this work, $g$ is given by the CBLiP architecture introduced in the next section. During training, we corrupt either the head or tail of a true triple ($p_i$) and obtain a corrupted triple ($n_i$). We then train the model to assign higher scores to true triples using a margin-based ranking loss:
\[\mathcal{L} = \sum_{i=1}^{|\mathcal{F}_{\mathrm{train}}|}\max(0, g(n_i) - g(p_i) + \gamma)\]
Here, $\gamma$ is the margin that separates the true and corrupted facts and allows for flexibility in training. At test time, triples with higher scores are considered to be more plausible completions.

\subsection{Transductive Setting} In the transductive setting, there is no separate test graph and the test-time inference triples satisfy $F_{\mathrm{infer}}\subset(\mathcal{E}_{\mathrm{train}}\times\mathcal{R}\times\mathcal{E}_{\mathrm{train}})$. That is, the entities seen at test time are all present in the training graph. Our goal in relation prediction is to, given putative head and tail entities, predict the relation between them. There are typically many fewer relations than entities (a few hundred vs. tens of thousands), so instead of a scoring function, we explicitly model the likelihood of each relation given the putative head and tail entities:
\begin{equation}
    \mathbb{P} (r| h, t) \propto g(h,t, \mathcal{N}_h, \mathcal{N}_t)
    \label{eqn:model}
\end{equation}
One advantage of explicitly modeling $\mathbb{P} (r| h, t)$ is that there is no need for negative samples.
The model is trained by minimizing the cross-entropy loss between the log-likelihood of our estimation and the ground truth observation $\mathbf{r}$ over the training data. CBLiP is again used as the architecture for $g$ in this setting.

\section{Model Overview}

This section describes the architecture used for $g$ in the CBLiP model. Given $\langle h,r,t\rangle$, we find the neighboring triples of both $h$ and $t$ as the context of that triple. In the inductive setting, each triple in our model is represented using learned entity role vectors for its \textit{head} and \textit{tail} entity along with a \textit{relation} vector embedding. The resulting vector encodings of the triples in the neighborhood are used as inputs to a connection-biased Transformer, and the output sequence is passed through a linear transformation to obtain the final score. Below, we explain these components in detail and the modifications made for the transductive setting. 

\subsection{Entity Role Vectors}
Inductive learning is aided by learning entity behaviors and interactions to facilitate inference on the unseen (during training) entities during test time. To this end, we represent each unseen entity by a vector representing its role. Traditional GNN-based models \cite{teru2020inductive, chen2021topology, lin2022incorporating} distinguish neighbor entities in a subgraph by assigning relative distance from the target \textit{head} or \textit{tail} entity and by initializing the values with one-hot vector encoding. 

Instead, we represent entities in the local neighborhood of the putative triple of interest by assigning roles to them. Such an entity has one of three roles: \{\textit{head, tail,} and \textit{other}\}. The role simply represents whether the neighbor entity is in fact the putative \textit{head} entity, the putative \textit{tail} entity, or some other entity. Despite the simplicity of the role embeddings, this distinction allows the model to distinguish between triples that are immediate neighbors or distant neighbors of the putative triple, thus improving the model's performance. This approach also ensures a shared representation of these roles across the model, unlike the local updates of relative-distance-based labeling. We denote the role vector of an entity $e$ succinctly with $\text{ROLE}(e)$.

\subsection{Context Embedding and Representation of a Triple}
We employ a connection-biased Transformer Encoder to obtain contextual embeddings for a given target triple $\langle h,r,t\rangle$.  We define the neighborhood of $h$ and $t$ as:
\[ \mathcal{N} = \{f| f\in\mathcal{N}_h\oplus\mathcal{N}_t\} \]
and use a breadth-first search algorithm to collect the triples in $\mathcal{N}$. Here $\oplus$ is the union or intersection operation. We select up to $m$ neighboring triples from $\mathcal{N}$; here $m$ is a hyperparameter. We obtain the embedding of each triple $f=\langle e_1,r,e_2\rangle$ in $\mathcal{N}$ by aggregating its entity and relation embeddings: 
\[\mathbf{f} = \mathcal{A}\big(\text{ROLE}(e_1), \mathbf{r}, \text{ROLE}(e_2)\big)\]
We explored two options for the aggregation function $\mathcal{A}$: concatenation and mean. The connection-biased aspect of the Transformer Encoder is described in the following section.

\subsection{Input Sequence}
With different pieces of information at hand, we can construct the final contextual representation of the target triple $\langle h, r, t\rangle$ for scoring. For each triple in $\mathcal{T}_{ \mathrm{train}}$, we construct a sequence of tokens encoding its neighborhood, $\mathcal{S}_{in}$:
\begin{equation}\label{eq:ind_seq}
    \mathcal{S}_{in} = [\mathbf{f}^\star, \mathbf{f}^1_{\mathcal{N}},
    \dots\mathbf{f}^m_{\mathcal{N}}]
\end{equation}
Here, $\mathbf{f}^\star$ is the embedding of the putative completed triple to be scored. This embedding is additionally aggregated (using $\oplus$) with a special \verb|target| vector embedding to distinguish it from the embeddings for triples from $\mathcal{N}$.

\subsection{Connection-Biased Adjacency Matrix}

Paths are often used to learn relation interaction patterns in a (sub)graph. We want to avoid the design decisions, time, and memory complexity that come with this addition to a model. By constructing a connection-biased adjacency matrix, we aim to learn implicit knowledge of paths, distance, and shared neighborhoods, which is instrumental in correctly predicting the final relation.

While GRAN \cite{wang2021link} constructs a similar matrix for each of its $n$-ary fact's components, we build this for members in a subgraph. The connection types in our model depict the overlap of entities between neighboring triples whereas GRAN distinguishes connections between $n$-ary fact components such as entity-value, attribute-value, and so on.

We construct a connection-biased adjacency matrix $\mathbf{C}$ for the triples in a subgraph. We do so by comparing whether these triples share the same head and tail entity and in which manner. Each entry $c_{ij}\in\mathbf{C}$ represents the kind of connections two triples $f_i$ and $f_j$ can have:

\begin{enumerate}
    \item $f_i$'s head is $f_j$'s head
    \item $f_i$'s tail is $f_j$'s tail
    \item $f_i$'s tail is $f_j$'s head
    \item $f_i$'s head is $f_j$'s tail
    \item $f_i$'s head is $f_j$'s head AND $f_i$'s tail is $f_j$'s tail (implying there are two parallel edges between the same pair of entities)
    \item $f_i$'s head is $f_j$'s tail AND $f_i$'s tail is $f_j$'s head (inverse relations, e.g., \verb|sonOf| and \verb|motherOf|)
    \item $f_i$ and $f_j$ share neither head nor tail
\end{enumerate}
\begin{figure}[t]
 \centering
\includegraphics[width=\linewidth]{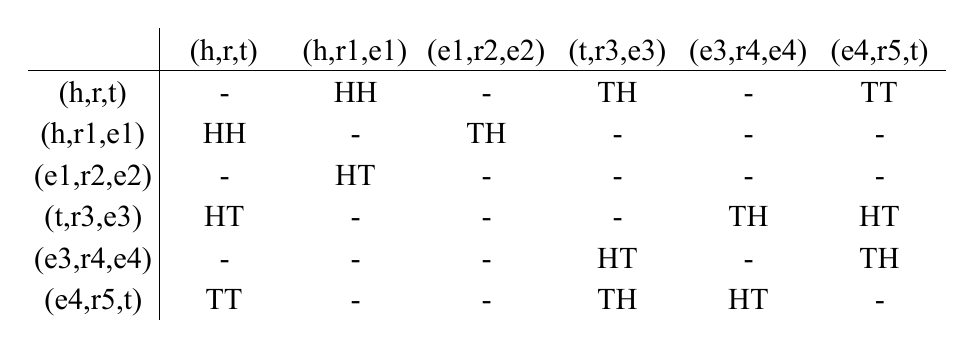}

\caption{An example of constructing a connection-biased adjacency matrix. The non-empty cells denote the presence of a particular kind of overlap of entities between triples.}
\label{fig:connection_}
\end{figure}
This approach serves three main purposes:
\begin{enumerate}
     \item It informs the model whether the head and tail entities have shared neighbor entities. 
     \item It serves as an implicit method of encoding paths as we just need to know about shared entities, and their order in a path. For example, we can have a pair of length-2 paths where the involved entities and relations are the same but their directions are different as follows:
\begin{itemize}
    \item $\langle e_1,r_1,e_2\rangle$ and $\langle e_2,r_2,e_3\rangle$
    \item $\langle e_1,r_1,e_2\rangle$ and $\langle e_3,r_2,e_2\rangle$
\end{itemize}
Our approach can distinguish between these two and any other combinations of directions.
    \item It implicitly captures the relative distance between all entity pairs. Most importantly, it informs the model of 1-hop neighbors and neighbors farther hops away from the target head entity (or from the tail entity).
   
\end{enumerate}
This implicit knowledge of paths, distance, and shared neighborhood is instrumental to correctly predicting the final relation. Figure \ref{fig:connection_} shows an example of finding four kinds of connections.

\subsection{Connection-Biased Attention in Transformer Encoder}

Transformers create key, query, and value vectors $\mathbf{K}, \mathbf{Q}, \mathbf{V}$ for tokens by a linear transformation with corresponding learnable weight matrices $\mathbf{W}^K, \mathbf{W}^Q, \mathbf{W}^V$ \cite{vaswani2017attention}. With slight abuse of notation, we represent any two tokens in an input sequence at $x_i$ and $x_j$. We define the connection-biased similarity between $x_i$ and $x_j$ as:


\begin{equation}
    \alpha_{ij} = \frac{(\mathbf{W}^Q\mathbf{x}_i)^\top(\mathbf{W}^K\mathbf{x}_j+\textbf{c}^K_{ij})}{d_y}
\end{equation}
Here, $\mathbf{c}_{ij}^K$ is the Key-specific bias vector for connection type $c_{ij}$. The corresponding output vector $ \textbf{y}_i$ is computed by modifying the attention computation:
\begin{equation}\label{eq:eb}
    \textbf{y}_i = \sum_{j=1}^{m+1}\frac{\exp(\alpha_{ij})}{\sum_{k=1}^{m+1}\exp(\alpha_{ik})}(\mathbf{W}^V\mathbf{x}_j+\mathbf{c}^V_{ij})
\end{equation}
Here, $\mathbf{c}_{ij}^V$ is the Value-specific bias vector for connection type $c_{ij}$.

\begin{figure}[t]
 \centering
\includegraphics[scale=0.5]{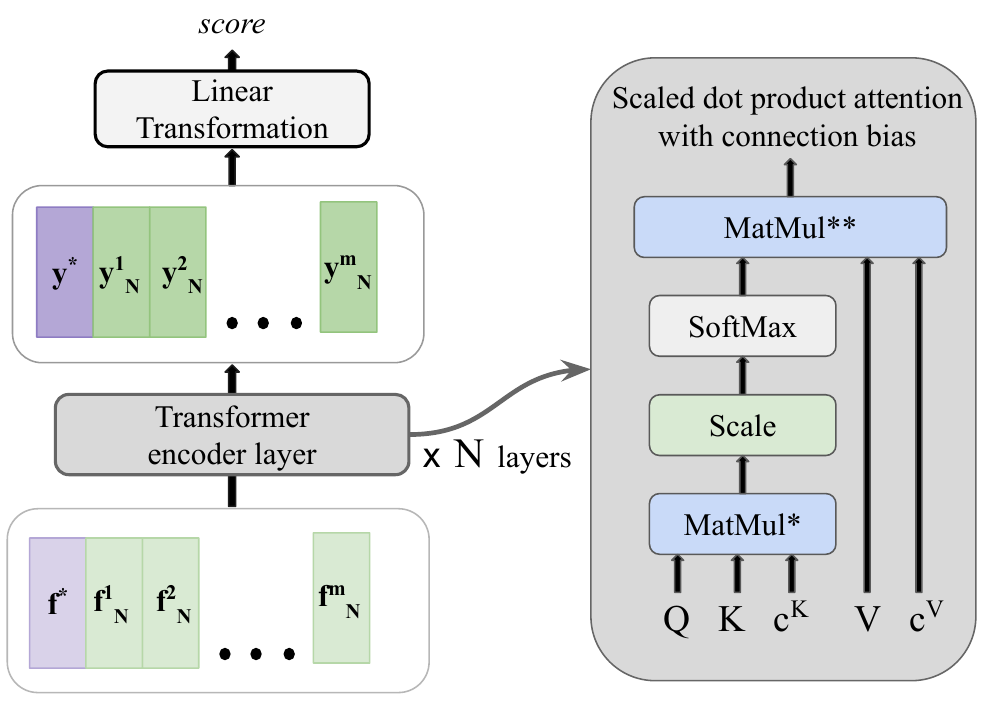}
\caption{Connection-biased attention computation. The left diagram shows an overview of the input and output sequence, and the right one elaborates the enhanced encoder layer with connection bias. Here, MatMul* and MatMul** refers to the modified matrix multiplication with key bias and value bias, respectively.}
\label{fig:eb_model}
\end{figure}

We denote the output sequence as:
\begin{equation}
    \mathcal{S}_{out} = 
    [\mathbf{y}^\star, \mathbf{y}^1_{\mathcal{N}},
    \dots,\mathbf{y}^m_{\mathcal{N}}]
\end{equation}
The architecture for connection-biased adjacency and the overall architecture of the proposed model is depicted in Figure \ref{fig:eb_model}.
\subsection{Transductive Training}

For using our model in the transductive training, we replace $\text{ROLE}(e)$ with entity-specific learnable vector embeddings. Note that, in this setting, all test entities are seen during training, and a relative representation is not needed. Another change in the transductive version of the model is the distinction between neighbors of \textit{head} and \textit{tail}. We choose $m$ neighbors from each entity and construct a specific interpretation of the input token sequence in Eq. (\ref{eq:ind_seq}).

\begin{table*}[t]

\begin{center}
\begin{tabular} { p{2cm}p{1.4cm}|p{0.7cm}p{0.7cm}p{0.7cm}p{0.7cm}|p{0.7cm}p{0.7cm}p{0.7cm}p{0.7cm}|p{0.7cm}p{0.7cm}p{0.7cm}p{0.7cm} }

\hline
 && \multicolumn{4}{c|}{WN18RR} &\multicolumn{4}{c|}{FB15K-237}   & \multicolumn{4}{c}{NELL995}\\
Methods && v1 & v2 & v3 & v4& v1 & v2 & v3 & v4& v1 & v2 & v3 & v4\\ \hline

&NeuralLP & 74.37	&68.93	&46.18&	67.13&	52.92&	58.94&	52.90&	55.88&	40.78&	78.73&	82.71&	80.58\\
Rule-based&DRUM &74.37&68.93&46.18&67.13&52.92&58.73&52.90&55.88&19.42&78.55&82.71&80.58\\
&RuleN &80.85&78.23&53.39&71.59&49.76&77.82&87.69&85.60&53.50&81.75&77.26&61.35\\
\hline
\multirow{3}{2cm}{Graph-based (w/o path)} &GraIL &82.45&78.68&58.43&73.41&64.15&81.80&82.83&89.29&59.50&93.25&91.41&73.19\\
&CoMPILE &83.60&79.82&60.69&75.49&67.64&82.98&84.67&87.44&58.38&\underline{93.87}&92.77&75.19\\
&TACT* & 84.04&81.63&67.97&76.56&65.76&83.56&85.20&88.31&79.80&88.97&94.02&73.78\\
\hline
\multirow{3}{2cm}{Graph-based (with path)}&
ConGLR &85.64&\underline{92.93}&70.74&\underline{92.90}&68.29&85.98&88.61&89.31&\underline{81.07}&\textbf{94.92}&94.36&\underline{81.61}\\
&Report &88.03&85.83&72.31&81.46&71.69&88.91&\underline{91.62}&\underline{92.28}&-&-&-&-\\
&LogCo &90.16&86.73&68.68&79.08&73.90&84.21&86.47&89.22&61.75&93.48&\underline{94.44}&80.82\\
&NBFNet &\underline{94.80}&90.50&\textbf{89.30}&89.00&\underline{83.40}&\textbf{94.90}&\textbf{95.10}&\textbf{96.00}&-&-&-&-\\
\hline
&\textbf{CBLiP} & 
\textbf{97.30}&\textbf{94.10}&\underline{81.30}&\textbf{96.40}&\textbf{89.30}&\underline{94.10}&84.20&80.10&\textbf{88.00}&93.70&\textbf{97.70}&\textbf{87.60}\\
\hline
\end{tabular}
\caption{Hits@10 for entity prediction in inductive KG dataset splits; Bold and underlined text represents the best and 2nd best results, respectively. All results are sourced from the original papers except for TACT (taken from ConGLR).}
\label{tab:Hits@10}     
\end{center}
\end{table*}

Note that in this case, $\mathbf{f}^\star$ is the target entity pair representation without the relation. Two separate vector embeddings specify the roles of each triple (neighbor of \textit{head} or neighbor of \textit{tail}) in the neighborhood. All respective equations are modified to reflect the presence of $2m+1$ triples in the input sequence. We make a final change by applying the softmax function to the output of the linear transformation to get a probability distribution over all possible relations.

\section{Experimental Evaluation}
		
\begin{table*}[t]

\begin{center}
\begin{tabular} { p{2cm}p{1.4cm}|p{0.7cm}p{0.7cm}p{0.7cm}p{0.7cm}|p{0.7cm}p{0.7cm}p{0.7cm}p{0.7cm}}

\hline
 && \multicolumn{4}{c|}{WN18RR} &\multicolumn{4}{c}{FB15K-237}  \\
Methods && v1 & v2 & v3 & v4& v1 & v2 & v3 & v4\\ \hline

\multirow{3}{2cm}{Rule-based}&NeuralLP&71.74&68.54&44.23&67.14&46.13&51.85&48.7&49.54\\
&DRUM&72.46&68.82&44.96&67.27&47.55&52.78&49.64&50.43\\
&RuleN&79.15&77.82&51.53&71.65&45.97&59.08&\textbf{73.68}&\textbf{74.19}\\
\hline
\multirow{2}{2cm}{Graph-based (w/o path)}&GraIL&80.45&78.13&54.11&73.84&48.56&62.54&70.35&70.6\\
&CoMPILE&78.28&79.61&53.97&75.34&50.52&\underline{64.54}&66.95&63.69\\
\hline
{Graph-based (with path)}&Report&\underline{80.95}&\underline{82.01}&\underline{58.38}&\underline{77.34}&\underline{53.22}&\textbf{70.62}&\underline{71.51}&\underline{71.28}\\
\hline
&\textbf{CBLiP} & \textbf{87.70}&\textbf{87.00}&	\textbf{60.50}	&\textbf{88.00} & 	\textbf{59.30	}&	63.70	&56.10	&	53.40\\
\hline
\end{tabular}
\caption{MRR for entity prediction in inductive KG dataset splits; Bold and underlined text represent best results and 2nd best results, respectively.}
\label{tab:mrr} 
\end{center}
\end{table*}
In this section, we evaluate our model on the entity prediction task in the inductive setting and present the performances in three KG datasets (12 versions). Additionally, we present relation prediction results in the transductive setting. All experiments were conducted on Quadro RTX 6000 (with NVLink) GPU with 32 GB memory. 
\subsection{Inductive Entity Prediction}
\subsubsection{Baselines}
NeuralLP \cite{yang2017differentiable}, DRUM \cite{sadeghian2019drum}, and RuleN \cite{meilicke2018fine} are rule-extraction-based methods. GraIL \cite{teru2020inductive}, CoMPILE \cite{mai2021communicative} and TACT \cite{chen2021topology} are graph-based models that use only the subgraph information surrounding a target entity pair and exclude explicit path information. ConGLR \cite{lin2022incorporating}, Report \cite{liu2023inductive}, LogCo \cite{pan2022inductive}, NBFNet \cite{zhu2021neural} are graph-based models that utilize path information between target \textit{head} and \textit{tail} entities. Report \cite{li2023inductive} uses vanilla Transformers to encode context and path and is the most similar to our model in terms of choice of encoding architecture.

\begin{table*}[t]
\begin{center}

\begin{tabular} { p{2.4cm}p{1.4cm}|p{.9cm}p{.9cm}p{.9cm}|p{.9cm}p{.9cm}p{.9cm}|p{.9cm}p{.9cm}p{.9cm}}

\hline
 & &\multicolumn{3}{c|}{WN18RR}  & \multicolumn{3}{c|}{FB15K-237} & \multicolumn{3}{c}{NELL995}\\ 

& Methods & MRR & Hits@1 & Hits@3 & MRR & Hits@1 & Hits@3 & MRR & Hits@1 & Hits@3 \\
\hline
&TransE &  0.784 & 0.669 & 0.870 & 0.966 & 0.946 & 0.984 & 0.841 & 0.781 & 0.889 \\
&ComplEx &  0.840 & 0.777 & 0.880 & 0.924 & 0.879 & 0.970  & 0.703 & 0.625 & 0.765\\
\multirow{3}{*}{Models w/o path}&DistMult   & 0.847 & 0.787 & 0.891 & 0.875 & 0.806 & 0.936  & 0.634 & 0.524 & 0.720\\
&RotatE   & 0.799 & 0.735 & 0.823 & 0.970 & 0.951 & 0.980 & 0.729 & 0.691 & 0.756 \\
&SimplE   & 0.730 & 0.659 & 0.755 & 0.971 & 0.955 & 0.987 & 0.716 & 0.671 & 0.748 \\
&QuatE   & 0.823 & 0.767 & 0.852 & \underline{0.974} & \underline{0.958} & 0.988 & 0.752 & 0.706 & 0.783 \\
&DRUM   & 0.854 & 0.778 & 0.912& 0.959 & 0.905 & 0.958  & 0.715 & 0.640 & 0.740 \\

\hline
Model with path &PathCon   & 

\underline{0.974} & \underline{0.954} & \textbf{0.994} & \textbf{0.979}  & \textbf{0.964}  & \textbf{0.994}  & 
\underline{0.896} & \underline{0.844} & \underline{0.941} \\

\hline

&\textbf{CBLiP}   & 
\textbf{0.976} & \textbf{0.960} & \underline{0.993} &  
0.971 & 0.949 & \underline{0.992} & 
\textbf{0.919} & \textbf{0.868} & \textbf{0.964} \\
\hline
\end{tabular}
\caption{Relation prediction in transductive KG datasets; Bold and underlined text represents best and 2nd best results, respectively. All results are sourced from PathCon.}
\label{tab:relation_results_1} 
\end{center}
\end{table*}

\begin{table}
\setlength{\tabcolsep}{1mm}

\begin{center}
\begin{tabular} 
 {ccccc}
\hline

 & v1 & v2 & v3 & v4\\ 
\hline
&\multicolumn{4}{c}{Hits@10}\\
\hline

CBLiP&\textbf{97.30}&\textbf{94.10}&\textbf{81.30}&\textbf{96.40}\\
CBLiP-vanilla&92.00	&71.70	&69.30&	90.90\\
\hline
&\multicolumn{4}{c}{MRR}\\
\hline
CBLiP&\textbf{87.70}	&\textbf{87.00}&	\textbf{60.50}&	\textbf{88.00}\\
CBLiP-vanilla&76.70&	66.20&	53.50&	77.90\\
\hline
\end{tabular}
\caption{Hits@10 and MRR for entity prediction in Wordnet dataset splits; Bold text represents better results. CBLiP-vanilla denotes our model without the connection-biased attention component.}
\label{tab:ablation}  
\end{center}
\end{table}

\subsubsection{Datasets}
\cite{teru2020inductive} extracted 12 inductive datasets from three popular KG benchmark datasets -- Wordnet, Freebase, and Nell. We present the dataset statistics in supplementary material \cite{cblip_extended}.

\subsubsection{Experimental Setup}

For inductive entity prediction, we corrupt a triple by replacing its head/tail with a randomly chosen entity during training. For inference, we want to see how the model ranks variations of corrupted triples by scoring $\langle h,r,?\rangle$ or $\langle ?,r,t\rangle$ and finding the rank of the true triple. We follow the existing literature and use 50 randomly chosen entities from $\mathcal{E}_{\mathrm{test}}$ to corrupt test triples. The model scores the true triples and the corrupted triples, and the rank of the true triple is recorded. The ranks of all true test triples contribute to finding  Hits@$n$ (ratio of correct hits in the top $n$ sorted predictions), which is a commonly used metric in rank-based experimental studies. We report Hits@10 and mean reciprocal rank (MRR) and evaluate the performance of our model. The hyperparameter selection is described in the supplementary material \cite{cblip_extended}.

\subsubsection{Entity Prediction Results}
The results of the entity prediction task according to Hits@10 are presented in Table \ref{tab:Hits@10}. Our model CBLiP achieves state-of-the-art performance in 7 dataset splits and the performance is consistently better than the rule-extraction and no-path baselines. The model struggles in the Freebase dataset, where a path-based model best utilizes the extremely rich data density. The MRR results of this experiment are presented in Table \ref{tab:mrr}, where we notice a similar trend of dominating performance in Wordnet and not in Freebase.  

\subsection{Transductive Relation Prediction}
The transductive entity prediction task is expensive due to the full vocabulary testing. It is also interchangeable with relation prediction task \cite{wang2021relational}. We choose relation prediction as it is a faster measure of how well a model can learn to reason over a KG in the transductive setting. 

We remove the target relation $r$ from all triples and generate an input sequence $S_{in}$. The output probabilities of each relation type are sorted and the position of true relation $r$ is retrieved. This serves as the rank of the relation. We report Hits@1, Hits@3, and MRR for this task in Table \ref{tab:relation_results_1}.
We demonstrate the results on three primary datasets: WN18RR, FB15K-237, and NELL995. Our supplementary material \cite{cblip_extended} contains the dataset statistics, hyperparameter configuration, and additional experiments on other datasets. 

We compare CBLiP with a neural baseline PathCon \cite{wang2021relational} that uses path information explicitly in the model. Other baseline models include a variety of translation-based, factorization-based, and rule-extraction-based models. 

Table \ref{tab:relation_results_1} shows the relation prediction results. We see again that CBLiP performs better than all models that explicitly do not utilize paths. Similarly, it achieves better or more competitive performance against PathCon.

\subsection{Ablation Studies}
We study the effectiveness of the proposed attention by comparing it with vanilla attention in Transformers. We call this variation CBLiP-vanilla. We run these experiments on the Wordnet dataset and report the results in Table \ref{tab:ablation}. We observe that the performance of the model in both metrics drops significantly when we eliminate the connection bias from the transformer encoder.

\subsection{Discussion}
We notice that CBLiP performs well in Wordnet and Nell across settings. This could be due to the dense degree of Freebase (on average, each entity has a large number of immediate neighbors), which a path-based model could best utilize.

However, we argue that omitting the use of paths has its strengths. Firstly, computing paths between target pairs on the fly is highly expensive and researchers often fix a path length and precompute a set of paths between target pairs of test set. However, this can be a limitation in real applications, as one will come across a new target entity pair during test time and has to compute paths between them. This also comes with an extra set of hyperparameter tuning regarding path length, how many paths to use, how to represent inverse relations, and how to aggregate each path.

In contrast to these, our proposed method adds only a handful of new learnable bias vectors for each connection type between tokens. Adding the connection bias to the Transformer encoder block does not add significant computational overhead since these bias computations do not require additional matrix multiplications. The scaled dot product attention module (that we modify by adding connection-bias vectors) retains its complexity of $\mathcal{O}(N^2d)$ for a sequence of $N$ tokens with $d$ feature dimension.

The entity role introduced in our paper is also an intuitive construct and adds negligible overhead to represent all entities in the model.  Since these two constructs capture the similarity of involved entities, the model implicitly learns relative distance information along with possible path information between tokens.

\section{Conclusion and Future Work}
We propose CBLiP, a KG link prediction model using inexpensive and intuitive use of entity role and connection-bias in a subgraph. We show the effectiveness of our model in two KG settings in different tasks where CBLiP showcases excellent performance while being intuitive and relatively simpler to its contemporaries. Future work can address the fully inductive setting, where entities \emph{and} relations may be seen at test time that were not present at training time. 


\section{A. Codebase}
We provide the pytorch implementation of our code and hyperparameter configurations for all our experiments at: \url{https://github.com/shoron-dutta/CBLiP}.
\section{B. Inductive Entity Prediction}
\subsection{B1. Hyperparameter Selection}
We select at most $m$ neighbors from $k$-hop subgraphs of entities. We include the closest neighbors first and then allow for neighbors farther away. We select feature dimension $d$ from the set $\{20, 32, 40, 64, 80, 128\}$.  We use multi-headed attention with the number of attention heads $\{2,4\}$ and select the number of encoder layers from $\{2,3,4\}$. We use Adam optimizer with learning rate from the set $\{0.01, 0.001, 0.008, 0.0005\}$. The specific hyperparameter setting for each dataset can be found in our GitHub repository.

\subsection{B2. Dataset}
The inductive dataset splits provided by \cite{teru2020inductive} are widely used for the inductive KG completion tasks. We present the dataset statistics in Table \ref{tab:ind_dataset}.

\begin{table}[t]
\setlength{\tabcolsep}{1mm}
\centering
\caption{Statistics of datasets used in inductive link prediction experiments}
\label{tab:ind_dataset}

\caption{WN18RR}
\begin{tabular}{c|c|cccc}
\hline
\textbf{Version} & \textbf{Split} & $|\mathcal{R}|$ & $|\mathcal{E}|$ & \#TR1 & \#TR2 \\
\hline
v1 & train & 9 & 2746 & 5410 & 630 \\
   & test  & 9 & 922  & 1618 & 188 \\
\hline
v2 & train & 10 & 6954 & 15262 & 1838 \\
   & test  & 10 & 2923 & 4011  & 441 \\
\hline
v3 & train & 11 & 12078 & 25901 & 3097 \\
   & test  & 11 & 5084  & 6327  & 605  \\
\hline
v4 & train & 9  & 3861  & 7940  & 934  \\
   & test  & 9  & 7208  & 12334 & 1429 \\
\hline
\end{tabular}

\vspace{5mm}

\caption{FB15k-237}
\begin{tabular}{c|c|cccc}
\hline
\textbf{Version} & \textbf{Split} & $|\mathcal{R}|$ & $|\mathcal{E}|$ & \#TR1 & \#TR2 \\
\hline
v1 & train & 183 & 2000 & 4245 & 489  \\
   & test  & 146 & 1500 & 1993 & 205  \\
\hline
v2 & train & 203 & 3000 & 9739 & 1166 \\
   & test  & 176 & 2000 & 4145 & 478  \\
\hline
v3 & train & 218 & 4000 & 17986 & 2194 \\
   & test  & 187 & 3000 & 7406  & 865  \\
\hline
v4 & train & 222 & 5000 & 27203 & 3352 \\
   & test  & 204 & 3500 & 11714 & 1424 \\
\hline
\end{tabular}

\vspace{5mm}

\caption{NELL-995}
\begin{tabular}{c|c|cccc}
\hline
\textbf{Version} & \textbf{Split} & $|\mathcal{R}|$ & $|\mathcal{E}|$ & \#TR1 & \#TR2 \\
\hline
v1 & train & 14  & 10915 & 4687  & 414  \\
   & test  & 14  & 225   & 833   & 100  \\
\hline
v2 & train & 88  & 2564  & 8219  & 922  \\
   & test  & 79  & 4937  & 4586  & 476  \\
\hline
v3 & train & 142 & 4647  & 16393 & 1851 \\
   & test  & 122 & 4921  & 8048  & 809  \\
\hline
v4 & train & 77  & 4922  & 7546  & 876  \\
   & test  & 61  & 3294  & 7073  & 731  \\
\hline
\end{tabular}

\end{table}

\begin{table}[h]
\setlength{\tabcolsep}{1mm}
\caption{Statistics of transductive\\KG datasets} 
\label{tab:transductive_dataset}
\centering
\begin{tabular}{c|c|c}
\hline
 & WN18 & WN18RR \\
\hline
\#nodes & 40,943 & 40,943 \\
\#relations & 18 & 11 \\
\#training & 141,442 & 86,835 \\
\#validation & 5,000 & 2,824 \\
\#test & 5,000 & 2,924 \\
\hline
\end{tabular}

\vspace{3mm}

\begin{tabular}{c|c|c}
\hline
 & FB15K & FB15K-237 \\
\hline
\#nodes & 14,951 & 14,541 \\
\#relations & 1,345 & 237 \\
\#training & 483,142 & 272,115 \\
\#validation & 50,000 & 17,526 \\
\#test & 59,071 & 20,438 \\
\hline
\end{tabular}

\vspace{3mm}

\begin{tabular}{c|c|c}
\hline
 & NELL995 & DDB14 \\
\hline
\#nodes & 63,917 & 9,203 \\
\#relations & 198 & 14 \\
\#training & 137,465 & 36,561 \\
\#validation & 3,907 & 3,897 \\
\#test & 3,964 & 3,882 \\
\hline
\end{tabular}
\end{table}
The terms \#TR1 and \#TR2 refer to the following:
\begin{itemize}
    \item In train mode:
    \begin{itemize}
        \item \#TR1 refers to $\mathcal{F}_{\mathrm{train}}$, the set of triples used for training
        \item \#TR2 refers to $\mathcal{F}_{\mathrm{valid}}$, the set of triples we use for validation
    \end{itemize}
    \item In test mode:
    \begin{itemize}
        \item \#TR1 refers to $\mathcal{F}_{\mathrm{test}}$, the set of triples used as a fact graph (to collect topological and neighborhood data) for inference triples
    \item \#TR2 refers to $\mathcal{F}_{\mathrm{infer}}$, the set of triples to test the model's inference capabilities

    \end{itemize}
\end{itemize}

\begin{table}[h]
\caption{Transductive relation prediction in KG datasets}
\label{tab:transductive_results_shortened}
\centering
\begin{tabular}{l|ccc}
\hline
Methods & MRR & Hits@1 & Hits@3 \\
\hline
\multicolumn{4}{c}{FB15K} \\
\hline
TransE   & 0.962 & 0.940 & 0.982 \\
ComplEx  & 0.901 & 0.844 & 0.952 \\
DistMult & 0.661 & 0.439 & 0.868 \\
RotatE   & 0.979 & 0.967 & 0.986 \\
SimplE   & \underline{0.983} & \underline{0.972} & \underline{0.991} \\
QuatE    & \underline{0.983} & \underline{0.972} & \underline{0.991} \\
DRUM     & 0.945 & 0.945 & 0.978 \\
PathCon  & \textbf{0.984} & \textbf{0.974} & \textbf{0.995} \\
CBLiP    & 0.863 & 0.763 & 0.962 \\
\hline
\multicolumn{4}{c}{WN18} \\
\hline
TransE   & 0.971 & 0.955 & 0.984 \\
ComplEx  & 0.985 & 0.979 & 0.991 \\
DistMult & 0.786 & 0.584 & 0.987 \\
RotatE   & 0.984 & 0.979 & 0.986 \\
SimplE   & 0.972 & 0.964 & 0.976 \\
QuatE    & 0.981 & 0.975 & 0.983 \\
DRUM     & 0.969 & 0.956 & 0.980 \\
PathCon  & \textbf{0.993} & \textbf{0.988} & \textbf{0.998} \\
CBLiP    & \underline{0.991} & \underline{0.985} & \underline{0.996} \\
\hline
\multicolumn{4}{c}{DDB14} \\
\hline
TransE   & 0.966 & 0.948 & 0.980 \\
ComplEx  & 0.953 & 0.931 & 0.968 \\
DistMult & 0.927 & 0.886 & 0.961 \\
RotatE   & 0.953 & 0.934 & 0.964 \\
SimplE   & 0.924 & 0.892 & 0.948 \\
QuatE    & 0.946 & 0.922 & 0.962 \\
DRUM     & 0.958 & 0.930 & 0.987 \\
PathCon  & \underline{0.980} & \underline{0.966} & \underline{0.995} \\
CBLiP    & \textbf{0.981} & \textbf{0.967} & \textbf{0.995} \\
\hline
\end{tabular}
\end{table}

\section{C. Transductive Relation Prediction}
\subsection{C1. Datasets}
Initial KG experiments were done on Freebase and Wordnet datasets, FB15k and WN18, whose test sets contained inverse triples of training triples. It caused simpler models to perform well during test by memorizing training data. \cite{toutanova2015observed} proposed a corrected versions of these datasets- FB15K-237 and WN18RR, respectively. We have presented results on FB15k-237, WN18RR, and NELL995 in the main text of our paper.

PathCon \cite{wang2021relational} proposed DDB14 dataset which is generated from the disease database. The statistics of all 6 datasets are presented in Table \ref{tab:transductive_dataset}.

\subsection{C2. Experimental Evaluation}
Here, we present results in the other 3 datasets. Table \ref{tab:transductive_results_part_2} shows CBLiP's performance in comparison with the relation prediction baseline models. CBLiP performs achieves competitive performance in WN18 and DDB14. It struggles in FB15K. We think the high triples to entities ratio in this dataset could be the reason for this. The dense neighborhood information is best captured with a path-based model like PathCon \cite{wang2021relational}.

\bibliography{aaai25}

\end{document}